\documentclass[letterpaper, 10 pt, conference]{ieeeconf}  

\IEEEoverridecommandlockouts                              
\usepackage[utf8]{inputenc}
\usepackage[T1]{fontenc}
\usepackage{graphics} 
\usepackage{graphicx}
\usepackage{hyperref}
\usepackage{cite}
\usepackage{mathtools}
\usepackage[nolist,printonlyused]{acronym}
\usepackage{cleveref}
\usepackage{multirow}
\mathchardef\mhyphen="2D 

\usepackage{etoolbox}
\makeatletter
\patchcmd{\@makecaption}
  {\scshape}
  {}
  {}
  {}
\makeatother

\usepackage{todonotes}

\title{\LARGE \bf
	Generalizing to New Tasks via One-Shot Compositional Subgoals
}

\author{Bian Xihan*\thanks{*All authors are from the Centre for Vision, Speech and Signal Processing (CVSSP), University of Surrey, UK. {\tt \{x.bian, o.mendez, s.hadfield\}@surrey.ac.uk}} and Oscar Mendez* and Simon Hadfield*}

\begin{document}

	\maketitle
	\thispagestyle{empty}
	\pagestyle{empty}

	\begin{abstract}
		The ability to generalize to previously unseen tasks with little to no supervision is a key challenge in modern machine learning research. It is also a cornerstone of a future ``General AI''. Any artificially intelligent agent deployed in a real world application, must adapt on the fly to unknown environments.
		
		Researchers often rely on reinforcement and imitation learning to provide online adaptation to new tasks, through trial and error learning.
		However, this can be challenging for complex tasks which require many timesteps or large numbers of subtasks to complete. These ``long horizon'' tasks suffer from sample inefficiency and can require extremely long training times before the agent can learn to perform the necessary long-term planning.
		
		In this work, we introduce CASE which attempts to address these issues by training an Imitation Learning agent using
		adaptive ``near future" subgoals. These subgoals are re-calculated at each step using compositional arithmetic in a learned latent representation space.
		In addition to improving learning efficiency for standard long-term tasks, this approach also makes it possible to perform one-shot generalization to previously unseen tasks, given only a single reference trajectory for the task in a different environment.
		Our experiments show that the proposed approach consistently outperforms the previous state-of-the-art compositional Imitation Learning approach by 30\%.
		
	\end{abstract}

	\section{INTRODUCTION}
	
	As researchers seek to introduce robotic technology into various aspects of our society, the robots must be able to preform increasingly complex tasks with enhanced automation and generality. 
	These new tasks are often complex, with multiple implicit subgoals that vary depending on the environment. As such, it is common for only the target end goal to be specified explicitly. 
	For example, if we ask the robot to bring us a cup of coffee, the robot will need to know where we are, as well as where the kitchen is, the tools, and the procedure for making coffee. The effort of learning such a complex composite task is enormous. 
	More problematic is the fact that even if we provide explicit subgoal guidance: i.e. where our kitchen is, where the coffee machine is and how to use our coffee machine, this knowledge won't transfer to robots in other houses. Even for the individual robot the solution may be brittle, as simply moving the location of the coffee cups may cause the task to fail.
	
	The biggest learning challenge for complex tasks is the complexity itself.
	Any complex task would almost always requires a large number of steps to complete an episode. This is particularly true for tasks with terminal-only sparse rewards. The longer the average trajectory is, the broader we can expect an unbounded state-space to become, and the lower our sample efficiency will be.
	In an Imitation Learning setting, the use of expert trajectories helps alleviate the ``vanishing reward'' problem by providing feedback at each step of the trajectory. However, the exploration and  data efficiency problems remain.
	
		\begin{figure}[t]
	    \centering
	\includegraphics[width=\linewidth]{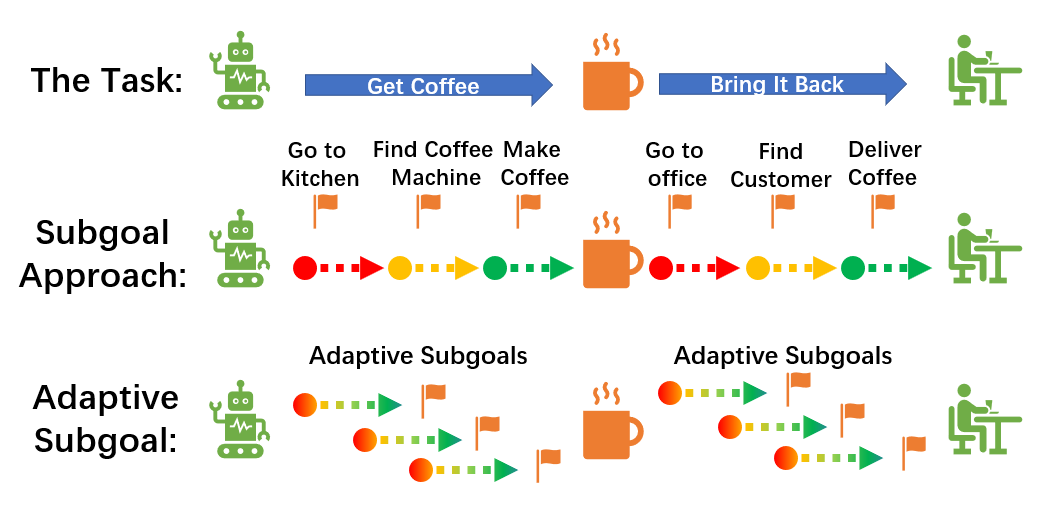}
	\vspace{-5mm}
	    \caption{In the ``make coffee and bring it back'' task, the traditional subgoal approach segments the complex task into smaller, manageable subtasks with clear end points and boundaries. In contrast the proposed CASE approach generates novel compositional subgoals which change at each step, gradually guiding the agent towards the final goal.}
	    \vspace{-5mm}
	    \label{fig:teaser}
	    
	\end{figure}

	The second challenge we seek to address is generalization. 
	In an Imitation Learning setting, the data efficiency challenge mentioned above will often manifest as a relatively restricted set of expert trajectories.
	As such learning to perform a complex task often involves repetitively training on a small set of sample tasks. This can easily lead to over-fitting on the training task set or the specific training examples of the tasks.
	A common approach to mitigate this, is to design the model hierarchically as shown in figure~\ref{fig:teaser}. In this case each stage of the model is intended to specialize in solving a certain class of problems. This can simplify  generalization within a subtask, but also exacerbates problems with data sparsity, as each submodel will only be exposed to a small portion of the training data.
	
	In contrast, our approach to solving these complex goal-oriented tasks, is to fully exploit the dataset to build a compositional task representation space, from which we can generate novel subgoals on the fly.
	We treat a single complex task as a sequence of smaller implicit tasks or subtasks. Each of these subtasks still requires multiple steps and some effort to learn. However, the need for long term planning (and the brittleness to divergences) is alleviated.
	Importantly, unlike previous approaches, we do not explicitly define a finite set of subtasks with hard boundaries (i.e. ``navigate to kitchen'', ``make coffee'' etc). Instead the subtasks can be any small sub-trajectory towards the overall goal (e.g. ``Move 3 meters towards the kitchen'', ``pick up the coffee beans'' etc) and is generated on-the-fly through compositionality. These sub-goals do not need to correspond to any task defined by the developer, nor do they need to have been previously observed during training. We refer to this approach as Compositional Adaptive Subgoal Estimation (CASE).

	Finally, an Imitation Learning policy is trained, using the learned compositional representation as it's state space, and with targets set via the adaptive subgoal estimation. These subgoals are adaptive as the rollout progresses, providing additional flexibility. Unlike traditional rigid and non-overlapping subgoals, our approach enables the agent to adapt to errors and drift, following alternative routes to the overall task, and avoiding deadlock with unachievable subtasks.
	In addition, the learning task is simplified as long term planning happens via the compositional space and the agent focuses on short term execution.
	This also allow the agent to perform one shot generalization over unseen tasks in the same environment. 
	With this approach, we are able to outperform standard imitation learning policies (including those using the same compositional state space) by over 30\% in unseen task generalization.

	In summary, the contributions of this paper are:
	\begin{enumerate}
		\item A novel approach estimate subgoal waypoints via a compositional task embedding space, referred to as CASE
		\item An Imitation Learning approach for complex compound tasks, based on online-subgoal estimation
		\item An evaluation of one-shot task-generalization for the policy, based on subgoal generalization
	\end{enumerate}

	\section{LITERATURE REVIEW}
	This section will first provide a broad overview of prior works on Subgoal Search from all areas of machine learning. It will then discuss specifically the current state-of-the-art in Imitation Learning, with a focus on compositional Imitation Learning approaches.
	
	\subsection{Subgoal Search}
	The most common use of Subgoal Search is in the context of path planning \cite{bengio2018advances}. Here, specific landmarks or semantic regions are recognized as a subgoal for the robot to navigate towards.
	These tasks often uses existing content within the observation as the subgoal, especially in visual related tasks \cite{gao2017intention, savinov2018semi, liu2020hallucinative,kim2021landmark }.
	
	More recently there has been significant research on extending Subgoal Search beyond landmark based navigation, particularly exploring the interaction with reinforcement learning (RL). 
	\cite{bakker2004hierarchical} was one of the earliest works combining hierarchical reinforcement learning with subgoal discovery. Here, high level policies discover subgoals while low level policies specialize in executing each subgoal.
	In a similar vein, \cite{kim2021landmark} trains a high-level policy with a reduced action space guided by landmarks.
	\cite{zhang2020generating} generalized this by looking for a sub-goal in a k-step adjacent region of the current state, within a general reinforcement learning environment.
	\cite{li2021active} builds upon this stable representation of subgoals, and proposes an active hierarchical exploration strategy which seeks out new promising subgoals with novelty and potential.
	All of these works discovered subgoals automatically, removing the need for hand-crafted subgoals. However, they all maintained the hard boundary between the subgoals.
	
	The majority of the research in subgoal estimation focuses on optimizing the generation or assessment of a candidate sub-goal state.
	However, \cite{czechowski2021subgoal} show a simple approach for efficiently generating subgoals which are precisely k-steps-ahead in reasoning tasks.
	In contrast our CASE approach chooses a possible ``near future'' subgoal, which adapts over time. This allows the policy network to focus on learning a more generalized skill for solving the task rather than attaining any specific subgoal. This in turn improves the performance of our technique when encountering unseen tasks, and provides robustness by allowing recovery from errors.
	
	\subsection{Imitation Learning}
	Imitation Learning (IL) approaches \cite{schaal1999imitation, silver2008high, chernova2009interactive, hester2018deep}
	utilize a dataset of expert demonstrations to guide the learning process.
	Initially, imitation learning relied heavily on supervision. 
	However, these approaches perform poorly with increasing episode length, and have issues with generalization. To improve upon this, the work of Levine, et al\cite{levine2013guided} uses expert trajectories to optimize the action policy rather than to imitate the expert exactly. 
	Later works move away from a passive collection of demonstrations. Instead they exploit the active collection of demonstration from expert. 
	To this end \cite{syed2007game, ho2016generative, song2018multi} create an interactive expert which provides demonstration as a response to the actions taken by the agent. This helps partially mitigate the issues with training data efficiency. 

	In more recent works, \cite{sun2017deeply, sun2018truncated} aim to combine IL with RL, by using IL as a pre-training step for RL.  \cite{cheng2018fast} perform randomized switching from IL to RL during policy training to enable faster learning.  \cite{murali2016tsc} uses expert trajectories to learn the reward function rather than the action policy. 
	
	To the best of our knowledge, the only prior work combining subgoal search and imitation learning is \cite{paul2019learning}. This approach uses a form of clustering on the expert trajectories to decompose the complex task into sub-goals. By learning to generate sub-goal states, the network obtains reward functions which direct the RL agent to move from one subgoal to another. However, the lack of compositionality in the model and the rigidity of the sub-goal prediction limit its generalization capability.
	
	\section{METHODOLOGY}
	
	Next, we detail our CASE method, utilizing adaptive subgoal waypoints for learning complex tasks and generalising to previously unseen tasks. First we clarify terminology.
	A ``task'' is defined as a singular goal the agent must complete through a series of interactions with the environment. A ``sequence'' of tasks means a collection of multiple separate tasks, some of which may be independent and some of which may depend on each other. Disregarding task dependencies, we allow the individual tasks within a sequence to be completed in any order, during the completion of the sequence.
	We further specify a ``complex task'' within this work, as a singular specified goal task, which nevertheless engenders an entire sequence of implicitly defined subtasks to be completed, due to the implicit dependencies within the environment.
	In our imitation learning framework, we define the ``subgoal waypoint'' as a state in the expert reference trajectory, located in the ``near future'' of the current agent's state. Note that the current trajectory and reference trajectory are both solving the same sequence of tasks, but are operating in different environments. Thus the subgoal waypoint cannot be used directly to guide the agent's trajectory.
	
	We learn a compositional latent space to represent tasks and sequences of tasks. 
	More specifically, a singular task maps to a unique point in the latent space. An unordered sequence of tasks (as defined above) also maps to a unique point in the latent space, which is the summation of the embeddings of all the subtasks within the sequence. This helps to draw a connection between ``complex tasks'' and ``task sequences'' as defined above. Both the singular complex task, and the explicit sequence of all dependent subtasks, should map to the same point within the latent space.
	This compositional approach makes manifest the lack of ordering specified above. The summation of subtask embeddings is an commutative operation, therefore changing the order of the summation does not change the final embedding.
	In order to learn this compositional task embedding, this constraint is codified as a number of regularization losses on the state encoder. 
	
	Finally we train agents to select actions using the learned task embedding, as their state representation. This provides a compact definition of both the current environment and the tasks to be completed.
	We tested our CASE framework for imitation learning. Taking the full set of states from timestep $t=0$ to the final timestep $t=T$, let $O_t$ be the observation of a state at time $t$ in a fully observable environment. We similarly specify $O_0^{ref}...O_T^{ref}$ as an expert reference trajectory which completes the same task sequence in a different environment. 
	The expert reference trajectories are extracted by greedy search over the environment for the optimal solution.
	\vspace{-2mm}
	\subsection{Compositional representation}
	A compositional representation is an embedding which encodes structural relationships between the items in the space \cite{mikolov2013distributed}.
	Consider a compositional representation $\vec{v}_{0,N}$ which encodes the trajectory and tasks required to progress from state $s_0$ to state $s_N$. More explicitly, $\vec{v}_{0,N}$ can be defined with a parameterized encoding function $g_{\phi}$ as:
	\begin{equation}
	    \vec{v}_{0,N}=g_{\phi}(s_0,s_N)=g_{\phi}(s_0,s_1)+...+g_{\phi}(s_{N-1},s_N),
	    \label{eq0}
	\end{equation}
	where $\phi$ are the encoding parameters.
	This representation defines a sequence of tasks as the sum of the representation for all subtasks within the sequence. 
	To prevent accidentally enforcing a specific ordering during the completion of these subtasks, the representation is built with commutativity, i.e. $\vec{v}_{A,B} + \vec{v}_{C,D} = \vec{v}_{C,D}+\vec{v}_{A,B}$
	This is a very powerful representation for computing encodings of implicit groups of subtasks. As an example, the embedding of all tasks that have yet to be accomplished at timestep $t$ a sequence can be calculated as $\vec{v} = \vec{u}_{0,N} - \vec{u}_{0,t}$.
	However, in a complex task sequence, the $\vec{v}$ often embeds a long trajectory which consist of many tasks. 
	This makes the learning process difficult, as information about far future tasks is a distraction from completing the current task.

	\subsection{Plan Arithmetic and Subgoal Waypoints}
	In one-shot imitation learning, the agent must perform a task (or sequence of tasks) conditioned on one reference example of the same task. In our work we further generalize this by allowing the current and reference task to be performed under different environments. 
	The agent is trained with many sequences of other tasks in other environments and then provided with an expert trajectory as reference to guide the new task, with no additional learning. Humans are adept at this: generalizing previous experiences to newly defined problems. However, for machine learning this is extremely challenging, and represents an important stepping stone towards general AI.
	
	During training, the agent is given two trajectories, the training trajectory $U$ and expert trajectory $R$ with matching task lists. It then learns a policy to perform online prediction of the actions in one trajectory, conditioned on the other trajectory as the reference. 
	In the running example `getting coffee', the agent will be provided with trajectories of retrieving coffee from a different office with a different floor layout.
	Learning how to make coffee without relying on specific meta knowledge about a particular environment is vital for improving generalization.

	To be more specific, a visual approach to task specification is taken. During both training and testing, the agent is given an image of the desired goal state for the current episode ($U_N$), as well as the goal state of the reference episode ($R_T$). It is also given an image of the current state ($U_t$), and an image of a future subgoal state ($R_I$) from the reference trajectory $\{R_0, R_T\}$.
	It is important to emphasise that the agent is not provided with any future knowledge about the current trajectory, beyond the target goal state which specifies the task to be completed. Subgoals are drawn from the future of the reference trajectory, not the current trajectory.
	
	To choose the subgoal waypoint, we assume the agent is always on the optimal path, therefore it's progress in the task is proportional to that the expert trajectory.
	As such when we choose the waypoint, we take the state $R_p$ in the reference trajectory, which has the same percentage of completion as in the training episode
	$\frac{p}{T} = \frac{t}{N}$, then add a fixed number $k$ steps to ensure the waypoint is in the ``near future" ($I=p+k$).
	The value of $k$ is determined experimentally in our work, as in previous works in the broader subgoal selection literature \cite{czechowski2021subgoal}.
	One potential issue with this approach is that the length of each subtask is unknown. If the current subtask in training episode is significantly longer or shorter than the expert trajectory, then the waypoint may fall into a different subtask. 
	However, we expect the agent to be able to adapt to this situation, as any state from the following subtask will already reflect the completion of the current subtask.
	
    Given our selected subgoal waypoint for this timestep, the model will first encode the compositional representation of the current state to the goal state ($\vec{U}=g_{\phi}(U_t,U_N)$). It will also encode the compositional representation of the reference sub-goal to the goal state of the reference episode ($\vec{V}=g_{\phi}(R_I,R_T)$).
	We can thus calculate a waypoint embedding $\vec{W}$ for the current trajectory with the following subtraction in the latent domain:
	\begin{equation} 
		\vec{W} = \vec{U} - \vec{V} = g_{\phi}(U_t, U_N) - g_{\phi}(R_I, R_T).
		\label{eq1}
	\end{equation}
	This estimates an embedded representation $\vec{W}$ of the trajectory from the current state ($t$) of the agent to the corresponding subgoal waypoint in the ``near future''.
	This short term task embedding is then used as the input for policy network $\pi \left(a_t|U_t, \vec{W}\right) $ to determine the actions of the agent.
	
	\subsection{Policy and encoder learning}
	The training of the compositional state encoder, and the subgoal based policy network, are undertaken concurrently.
	Both models are updated according to the policy loss:
	 \begin{equation}
		L_a\!(U_{t}, \!U_N,\! R_I, \!R_T\!) \!\!= \!\mhyphen log\!\left(\! \pi\!\left(\!\hat{a}_t|U_{t},g_{\phi}(U_t, \!U_N)\mhyphen g_{\phi}(R_I, \!R_T\right)\!\right)
		\label{eq2}
	 \end{equation}
	 where $\hat{a}_t$ is the expert reference action at the current state $U_t$.\looseness=-1
	
	Additionally, there are two regularization losses using a triplet margin loss function $l_m$\cite{chechik2010large}. The first ensures that arithmetic operations are correctly preserved within the latent space. To this end the compositionality is enforced by loss $L_H$ which ensures that the sum of the embedding of the past states and the embedding of the future states are equal to the embedding for the entire task 
	\begin{equation}
	L_H(U_0, U_t, U_N)\! =\! l_m(g_{\phi}({U_0, U_t}) + g_{\phi}({U_t, U_N}), g_{\phi}({U_0, U_N})).
	\label{eq3}
	\end{equation}
	The second regularization loss ensures that similarity in the latent space corresponds to semantically similar tasks. The full training and reference pair ($U, R$) should have similar embedded representations 
	\begin{equation}
	L_P(U_0, U_N, R_0, R_T) = l_m(g_{\phi}({U_0, U_N}), g_{\phi}({R_0, R_T})).
	\label{eq4}
	\end{equation}
	Thus the loss function for the framework is expressed as the weighted sum of the three losses: $L = L_a + \lambda_H L_H + \lambda_P L_P$.

	\section{EVALUATION}
	Our experiments aim to show the improvement of the agent's generalisation capability when trained using our CASE approach to online compositional subgoals.
	Therefore, we evaluate
	the performance on tasks which were previously unseen during training, and for which only a single reference episode is provided.
	In both cases, the observation is provided as a pair of images. The state encoder $g_{\phi}$ is a 4 layer CNN, and is shared by the reference and training episode. This encodes the current state to goal state sub-trajectory, as well as the sub-goal state to reference goal state sub-trajectory.
	The resulting latent will then be processed according to EQ1, and the output is fed into the policy network to estimate the action.
	
	In each experiment we contrast several variants of our own approach, including the effect of the current image branch and the additional compositionality losses. We also compare against the current state-of-the-art in compositional IL \cite{devin2019compositional}.
	Additionally, we include an ablation study on the ``near future'' subgoal lookahead parameter $k$. In all other experiments we set $k=4$. We also set the loss weightings $\lambda_H = \lambda_P = 1$.
	
	\subsection{Environment}
\begin{figure}[h]
	\vspace{-3mm}
	    \centering
	    \includegraphics[width=0.9\linewidth]{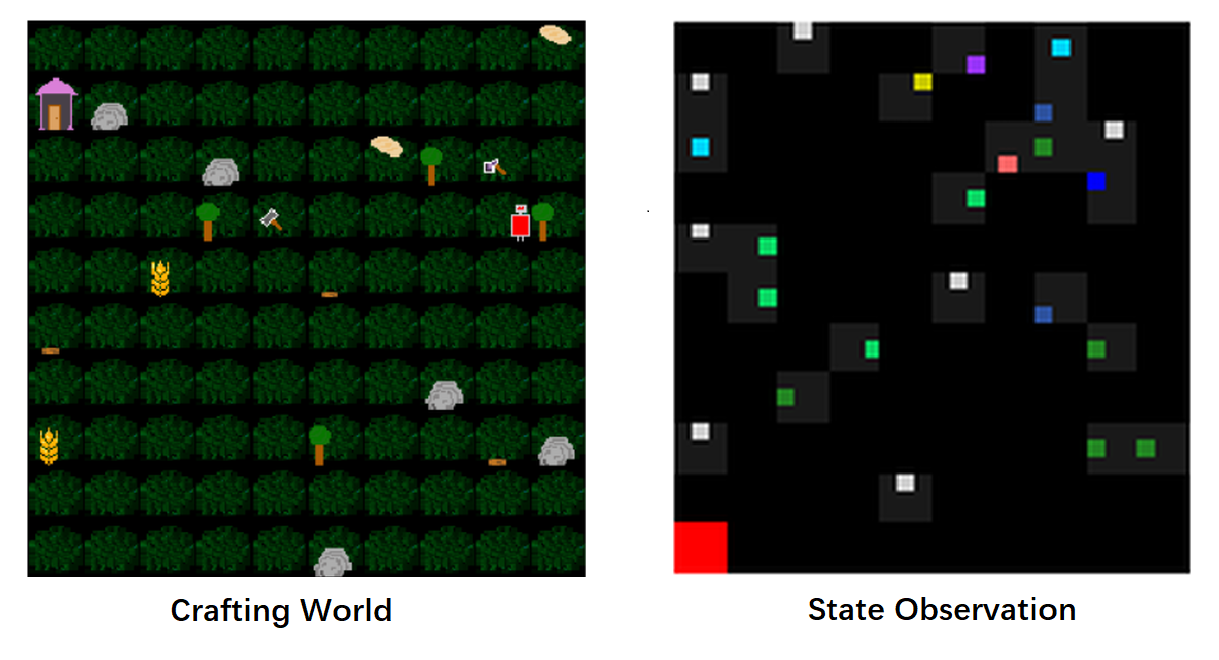}
	    \vspace{-2mm}
	    \caption{An example of the Craft World\cite{devin_craftingworld_2020} environment, on the left is a rendering of the environment, the right is an unrendered input state.}
	    \label{fig:data_example}
	\end{figure}

	In order to examine generalization and long-term IL problems, we trained our agent with the Craft World\cite{devin_craftingworld_2020} environment. This also facilitated fair comparison with the state-of-the-art compositional IL approach of \cite{devin2019compositional},
	This environment is a 2D discrete-action world with a top-down grid view where the agent is also able to move in one of 4 directions at each step. 
	There are different types of object in the environment as shown in figure~\ref{fig:data_example} including tree, rock, axe, wheat, bread, etc. 
	Some objects can be interacted with by an agent via pick up and drop off actions. 
	The object moves with the agent when it has been picked up, and can cause transformations to other objects in the environment. 
	For example, if the agent carries an axe to a tree, the tree will be transformed into a log, which then can be transformed into a house once the agent picks up a hammer and brings it to the log. 
	It is apparent that this environment, makes it possible to define complex long-horizon tasks such as ``make bread'' or ``build house'' which include many implicit subgoals. Furthermore, these tasks can be combined into sequences such as [``make bread'', ``eat bread'', ``build house'']. This provides a good selection of unique tasks to generate for training data. This also liberates the agent from skill list labels \cite{oh2017zero} or language based skill description \cite{shu2017hierarchical} which limits the generalisation to unseen tasks and sequences. 
	Also of interest is the fact that this allows the tasks sequence to be generated with no explicit ordering, giving more freedom in both data generation and generalization.
	
	The training and testing dataset are generated with a random map, upon which the agent is required to perform 2-8 different tasks in sequence with no particular ordering.
	The expert trajectory is generated with greedy search to ensure an optimal solution.
	The agent is trained on 150,000 episodes and tested on the same number of unseen episodes.
	To test one-shot generalization, the set of training tasks is different from the set of testing tasks, requiring generalization from the reference trajectory.

	\subsection{One-shot task generalization and ablation study}\label{sec:ablation}
	
	Results from the generalization test are shown in table \ref{tab:my-table}.
	The CASE agent is able to outperform the original SOTA benchmark \cite{devin2019compositional} by over 30\% in unseen task success rate. The additional assistive losses (CASE+L) only improved the performances by a small amount at later training steps.
	The SOTA (CPV-FULL) in it's original form \cite{devin2019compositional} does not take the desired goal state as input at test time. Instead the approach relies on the compositionality of the current and reference trajectory to produce a goal state. 
	Therefore we also evaluate and enhanced version of \cite{devin2019compositional} (CPV+ Goal Guidance), where the reference trajectory is replaced with the ground truth end-goal of the current trajectory at test time.
	As shown in table \ref{tab:my-table} the CASE approach is still able to outperform the modified SOTA consistently on unseen tasks. 
	
	

	
	The ablation study is performed over the different components of the framework. More specifically we explore the benefits of the assistive losses ($L_H$ \ref{eq3} and $L_P$ \ref{eq4}), and the current state image input ($U_t$).
	As shown in table \ref{tab:my-table}, the addition of both the current state image branch and the assistive losses each increased the performance of the model. 
	As expected, the current image provided additional information more relevant to the current task, while the assistive losses increased the generalization capability by encouraging a more regularized representation.
	
\begin{figure}
		\centering
		\includegraphics[width=0.75\linewidth]{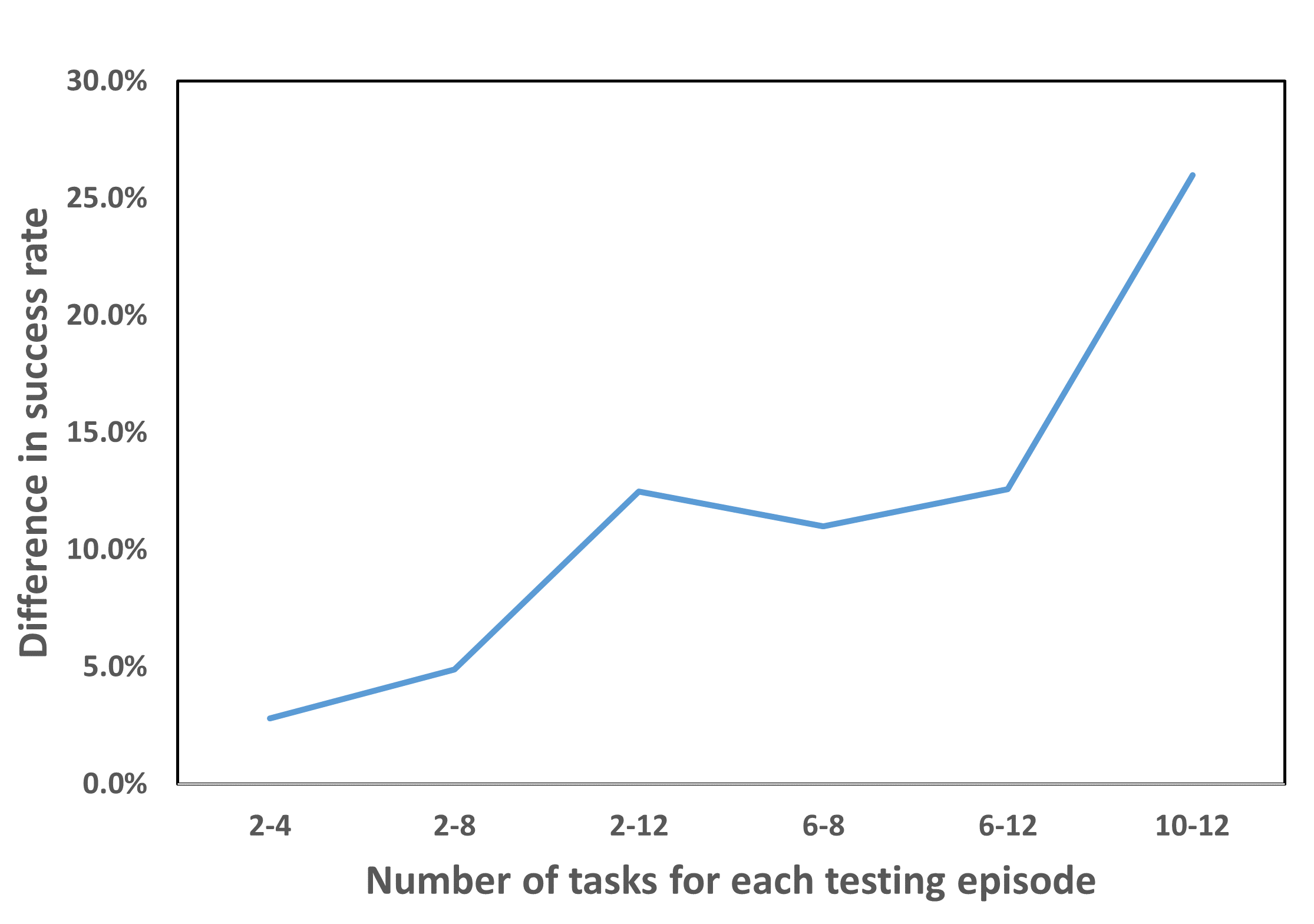}
		\vspace{-2mm}
		\caption{A graph of the difference in performance for CASE, and the original \& enhanced baselines, as the number of tasks in a sequence increases.
		}
		\vspace{-5mm}
		\label{fig:20220213result}
	\end{figure}

	During experimentation, we observed that the difference in performance between CASE and the enhanced SOTA grows as the number of tasks in a sequence increases.
	When the number of tasks in each sequence is chosen from the range 2-4, the CASE approach outperforms the baseline by around 3-4\%. When sequence lengths increase, the performance gap becomes much wider as shown in fig \ref{fig:20220213result}. It is worth mentioning that the CASE agent is able to maintain a success rate above 60\% throughout the experiments.
	
\begin{table}[h]
\centering
\begin{tabular}{llll}
Model &
  \begin{tabular}[c]{@{}l@{}}Best\\ Performance\end{tabular} &
  \begin{tabular}[c]{@{}l@{}}Average \\ Performance\end{tabular} &
  \begin{tabular}[c]{@{}l@{}}Standard\\ Deviation\end{tabular} \\ \hline
CPV-FULL\cite{devin2019compositional}                                                    & 0.432          & 0.392          & 0.0166 \\
\begin{tabular}[c]{@{}l@{}}CPV+Goal\\ Guidance\end{tabular} & 0.670          & 0.647          & 0.0146 \\
\textbf{CASE}                                                     & 0.689          & 0.641          & 0.0133 \\
\textbf{CASE+CI}                                                  & 0.701          & 0.676          & 0.0139 \\
\textbf{CASE+CI+L}                                                & \textbf{0.712} & \textbf{0.687} & 0.0167
\end{tabular}
\caption{{\footnotesize An ablation study on the different components of the network: Current state image (CI) and assistive losses (L).
}}
\vspace{-10mm}
\label{tab:my-table}
\end{table}

\begin{figure}[th]
		\centering
		\includegraphics[width=0.9\linewidth]{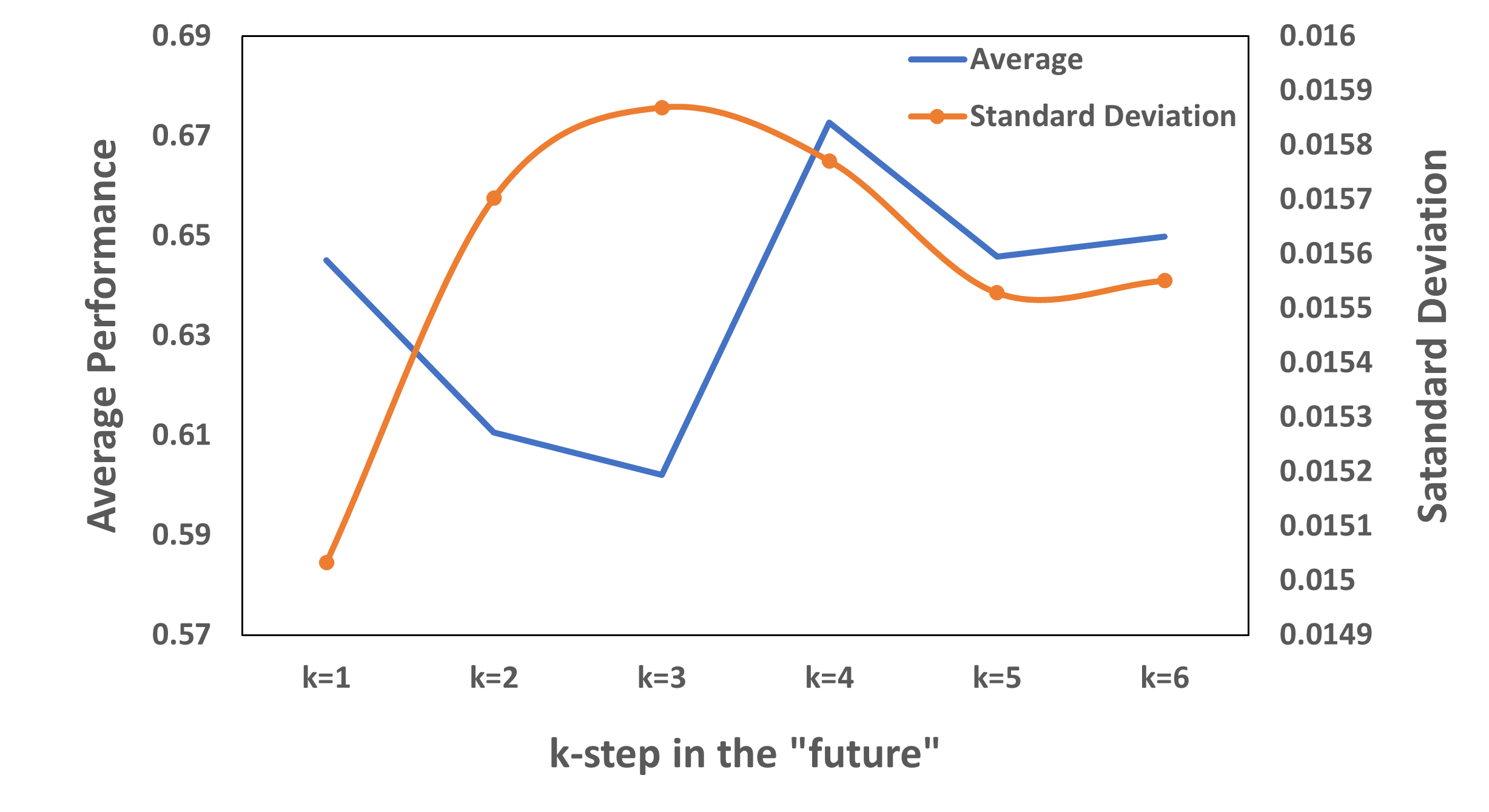}
			\vspace{-2mm}
		\caption{An ablation study over the $k$ parameter for choosing the ``near future'' waypoint. The average performance of each $k$-step is calculated from 8 trials.}
		\label{fig:k-step}
			\vspace{-6mm}
	\end{figure}

	Finally, we tested several settings for the ``near future'' lookahead parameter $k$. 
	As shown in figure \ref{fig:k-step}, when $k=4$ the agent's performance is maximized.
	The graph shows some sensitivity to the parameter $k$, and performance can be unstable at lower values. The setting $k=1$ has a middling performance and the lowest standard deviation, while $k=2$ and $k=3$ have the worst performance and highest variance.
	We suspect this is due to the inconsistency in the length of the randomized subtasks between the training episodes and expert trajectories.
	As an example, imagine the first task in the training episode is to pick up an axe. In the agent's training episode the axe may be 10 steps away from the current state, while it is 2 steps away in the expert reference trajectory. If the $k$ value is less than 2, then the generated subgoal will be after the completion of the current subtask within the expert trajectory. 
	The reverse can also be true for larger values of $k$ if the distances are swapped. 
	In most cases the agent is able to deal with this: a subgoal for the following task is still easier to learn from than the entire remaining trajectory. Nevertheless, it may be interesting for future work to explore the automatic computation of the optimal $k$ parameter during compositional subgoal estimation.
	
	\vspace{-3mm}

	\section{CONCLUSIONS}

	In this work, we proposed CASE, an approach to learn a compositional task representation which enabled novel subgoal estimation from reference trajectories in IL. This makes it significantly easier to learn long and complex sequences of tasks, including those with implicit or poorly defined subtasks.
	With this technique, we developed an IL agent which can generalize to previously unseen tasks with a success rate of around 70\%. This represents an improvement of around 30\% over the previous SOTA. 
	
	However, this approach can be developed further in future work.
	As discussed in section~\ref{sec:ablation}, using a fixed value for the $k$-step lookahead parameter may be suboptimal. Experiments indicate that performance and stability may be improved by developing an adaptive lookahead window, based on recent  developments in the broader field of subgoal search \cite{czechowski2021subgoal}.
	

	\addtolength{\textheight}{-7cm}   
    \section*{ACKNOWLEDGMENT}
	This work was partially supported by the UK Engineering and Physical Sciences Research Council (EPSRC) grant agreement EP/S035761/1 "Reflexive Robotics".
	
	\bibliographystyle{plain}
	\bibliography{PoseOpt.bib}
	
\end{document}